\pdfoutput=1

\documentclass[dvipsnames]{article}
\usepackage{arxiv}
\usepackage{graphicx}
\usepackage{amsmath}
\usepackage{amssymb}
\usepackage{booktabs}
\usepackage{multirow}
\usepackage{comment}
\usepackage[utf8]{inputenc}                                 %
\usepackage[australian]{babel}                              %
\usepackage[T1]{fontenc}                                    %

\usepackage[activate={true,nocompatibility},final,tracking=true,kerning=true,spacing=true,factor=1100,stretch=10,shrink=10]{microtype}
\microtypecontext{spacing=nonfrench}

\usepackage[inline]{enumitem}

\usepackage[autostyle=try,                      %
  style=australian]{csquotes}
  \MakeOuterQuote{"}                    %

\usepackage{amssymb}                                        %
\usepackage{amsmath}
\usepackage{braket}                                         %

\newcommand{\vect}[1]{\boldsymbol{\hat{#1}}}
\newcommand{\vectbar}[1]{\boldsymbol{\bar{#1}}}
\newcommand{\vectalt}[1]{\boldsymbol{#1}}
\usepackage{graphicx}                                       %

\usepackage{tikz}

\usepackage{subcaption} %

\usepackage{booktabs}
\usepackage{placeins}

\usepackage{xcolor}

\usepackage[boxed,ruled,vlined,linesnumbered]{algorithm2e}
\DontPrintSemicolon
\SetKwProg{Fn}{function}{}{}
\SetKwFunction{FnFeatureExtraction}{FeatureExtraction}
\SetKwFunction{FnPoiNet}{PoiNet}
\SetKwFunction{FnBoundingBoxGenerator}{BoundingBoxGenerator}
\SetKwFunction{FnActivityNet}{ActivityNet}
\SetKwFunction{FnSampleFree}{SampleFree}
\SetKwFunction{FnRestartArm}{RestartArm}
\SetKwFunction{FnPickArm}{PickArm}
\SetKwFunction{FnRewire}{Rewire}
\SetKwFunction{FnNearest}{Nearest}
\SetKwFunction{FnRestartArm}{RestartArm}
\SetKwComment{Comment}{$\triangleright$\ }{}
\SetKwInput{KwInit}{Initialise}
\SetKwInOut{KwPara}{Parameter}

\SetCommentSty{mycommfont}

\usepackage{cleveref}                                        %
\crefname{assumption}{assumption}{assumptions}
\crefname{problem}{problem}{problems}
\crefname{algorithm}{Alg.}{Algs.}
\Crefname{algorithm}{Algorithm}{Algorithms}
\crefname{figure}{Fig.}{Figs.} %
\crefformat{equation}{(#2#1#3)}
\crefrangeformat{equation}{(#3#1#4) to~(#5#2#6)}
\crefmultiformat{equation}{(#2#1#3)}%
{ and~(#2#1#3)}{, (#2#1#3)}{ and~(#2#1#3)}

\usepackage[
    backend=biber
  ,minnames=3                       %
  ,date=year                %
  ,maxbibnames=99
  ,labeldate=year
  ,style=authoryear
  ]{biblatex}
\AtEveryBibitem{                      %
  \iffieldequalstr{eprinttype}{jstor}
  {\clearfield{eprint}}
  {}
}

\DeclareSourcemap{
  \maps{
    \map{
      \pertype{inproceedings}
      \step[fieldset=publisher, null]
      \step[fieldset=urldate, null]
      \step[fieldset=volume, null]
      \step[fieldset=pages, null]
    }
  }
}
\renewrobustcmd*{\bibinitdelim}{\,} %
\newcommand{\poinet}{\textsc{p}\hspace*{-.08em}%
                     \textsc{o}\hspace*{-.08em}%
                     \textsc{i}{\smaller Net}%
                     \xspace}

\Crefname{figure}{Fig.}{Figs.}

\newcommand{\shortheadtitle}{Real-time Aerial Detection and Reasoning on Embedded-UAVs}

\begin{document}

\title{Real-time Aerial Detection and Reasoning on Embedded-UAVs in Rural Environments}

\author{Tin Lai%
\thanks{
\noindent Part of the work was done while Tin Lai was visiting the National Institute of Informatics, Tokyo 101-8430, Japan.
}%
\vspace{1mm}\\
{\tt\small oscar@tinyiu.com}\\
School of Computer Science\\
University of Sydney\\
Australia%
}

\maketitle

\begin{abstract}
We present a unified pipeline architecture for a real-time detection system on an embedded system for UAVs.
Neural architectures have been the industry standard for computer vision.
However, most existing works focus solely on concatenating deeper layers to achieve higher accuracy with run-time performance as the trade-off.
This pipeline of networks can exploit the domain-specific knowledge on aerial pedestrian detection and activity recognition for the emerging UAV applications of autonomous surveying and activity reporting.
In particular, our pipeline architectures operate in a time-sensitive manner, have high accuracy in detecting pedestrians from various aerial orientations, use a novel attention map for multi-activities recognition, and jointly refine its detection with temporal information.
Numerically, we demonstrate our model's accuracy and fast inference speed on embedded systems.
We empirically deployed our prototype hardware with full live feeds in a real-world open-field environment.
\end{abstract}

\section{Introduction}

Real-time awareness and understanding of the physical world have long been open questions for the robotics community.
For machines to successfully manoeuvre in the presence of dynamic obstacles, e.g. autonomous car driving within a crowd, one must perceive the intention of the humans and respond appropriately.
A real-time anomaly monitor system is crucial for crowd monitoring during mega-events~\autocite{schulte2017_AnalComb}, where the system must recognise any hostile intentions that any individuals might be planning to perform.
Moreover, an autonomous aerial surveying system in a search and rescue mission~\autocite{abdelkader2013_UAVBase} will require the ability to comprehend the state of any detected humans and reports accordingly~\autocite{hildmann2019using}; significantly when the number of dispatched drones greatly exceed the number of human operators~\autocite{aljehani2016multi}.

All the mentioned applications have a common interest---the ability to perform semantic understanding directly on a remote machine.
While some applications could be achieved with a centralised approach, i.e. perform semantic inferences on a centralised server farm, most existing infrastructure will have difficulties implementing such an approach.
For example, wireless bandwidth will be restricted for transferring video footage to a remote server from drones~\autocite{merino2006cooperative,sahingoz2013mobile,vega2015resilient}.
This is especially true during a search and rescue operation, where internet access might be unavailable, or the bandwidth is limited to reserve for critical communication.
Moreover, some privacy-sensitive applications might want to avoid sending sensitive information over the air to a server~\autocite{huang2019lightweight}.
Therefore, for specific applications, directly performing inferences on the embedded system of the remote hardware is the only approach.

\begin{figure}[t]
    \centering
    \includegraphics[width=.95\linewidth]{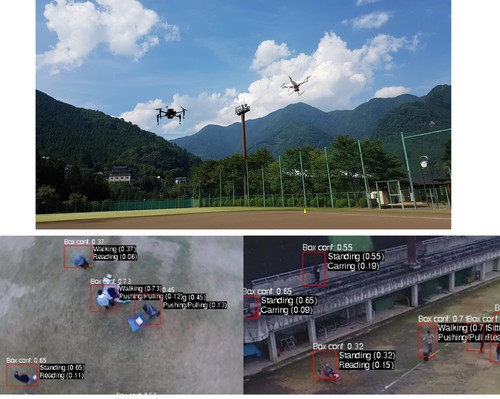}
    \caption{
        Deploying live detection and recognition in
        the field of \emph{Okutama, Japan} (top).
        Spatial locations of detected pedestrians, recognised actions and confidence scores are all processed on-board
        directly for transmitting to remote server.
        The processed information is overlaid on original video images for visualisation (bottom).
        \label{fig:intro-image}
        }
\end{figure}

The scope of this paper is on performing decentralised real-time autonomous activity reporting in a remote area, focusing on semantic understanding of people's current status from Unmanned Aerial Vehicles (UAVs) perspective.
UAVs play a crucial role when operating in rural environments, for tasks such as disaster response and recovery operations, where ground access is infeasible in the aftermath of natural disasters such as earthquakes.
It has been commonplace for mayors and governments to pursue mega-events to develop or regenerate their cities.
Mega-events, such as festive carnival, can bring positive socio-economic benefits to the city.
However, mega-events bring substantial strains to the security or police forces to monitor the environment, especially when human resources are scarce.
Similarly, in disastrous situations, autonomous UAVs surveying is hugely desirable due to the area of land they can cover.
Victims can be located faster and reached earlier when UAVs are deployed in mountain search and rescue operations~\autocite{karaca2018potential}.
However, current approaches transfer multi-gigabytes of video footage from remote drones to a centralised server for analysis.
Network activities are severely limited in disaster area~\autocite{chen2007real}, and any wireless bandwidth should be preserved for the communication channel between emergency services~\autocite{bai2010emergency} and critical communication.
Therefore, it severely limits relying on centralised servers for video processing.

On the other hand, lightweight models such as MobileNetV2~\autocite{sandler2018mobilenetv2} can achieve reasonable runtime speed but sacrifices predictive performance on aerial-captured images due to the highly varying camera angles from UAVs' perspective.
Although they are suitable for static embedded systems, the severely limited inference quality poses difficulty for usage on UAVs.
To this end, we propose our framework with the following properties.
We will employ UAVs with an embedded camera on board to survey an area of interest and report any abnormality to operators---namely, the current status of the detected pedestrians and associated GPS coordinates.
The problem setup requires a sizeable amount of inexpensive UAVs patrolling the vast aerial space for search and rescue; therefore, each UAV must operate autonomously and raise alerts if necessary.
Each UAV will be equipped with an embedded system and a pre-trained model for surveying.

This paper proposes a lightweight model pipeline that simultaneously detects pedestrians and recognises their current status.
Furthermore, our model is lightweight with minimal parameters and suitable for real-time processing in an onboard system.
A novel regression-based approach first performs preliminary human detection to address the camera perspective variance issue.
Then, the preliminary results are further refined by information from the framework's temporal layers to jointly recognise the detected pedestrians' current status.
Therefore, it uses minimal resources to detect abnormality directly on remote drones.
We deployed the live system at a remote field with real-time tracking and status reports delivered to the user through wireless transmission (\cref{fig:intro-image}) and provided comprehensive numerical results on the model performance.

\begin{algorithm}[!b]
    \caption{Overall model description}  \label{alg:overall}
    \While{Not terminated}{
        $\mathcal{I}_t \gets$ fetch latest image from camera at $t$ \;
        $\mathcal{F}_t \gets$ extract multiscale dense features from $\mathcal{I}_t$ \;
        $\mathcal{S}_t,\mathcal{R}_t \gets$ outputs from two separated convolutional layers with  $\mathcal{F}_t$ as input\;
        $\vect{b}_t \gets \FuncSty{BoxGenerator}(\mathcal{S}_t,\mathcal{R}_t)$\Comment*[r]{Alg.\ref{alg:box-generator}}
        $\vect{f}_t \gets $ crop $\mathcal{F}_t$ with each $b_t\in\vect{b}_t$\;
        associate each $f_{t}\in\vect{f}_t$ with the corresponding $f_{t-1}\in\vect{f}_{t-1}$ from previous time step\;
        generate $A_{t,i}$ and concat to each $f_{t}\in\vect{f}_t$\Comment*[r]{Eq.\ref{eq:attention-map-val}}
        $\vect{a}_t,\vect{c}_t\gets$ predict actions and confidence scores from the recurrent neural network with input $\vect{f}_{t}|\vect{f}_{t-1}$\;
        use $\vect{c}_t$ with non-maximum suppression on $\vect{b}_t$ and $\vect{a}_t$\;
        output $\vect{b}_t$, $\vect{a}_t$, $\vect{c}_t$ to user
    }
\end{algorithm}

\section{Related works}\label{sec:related-work}

Recent advancements in deep learning provide a robust framework for developing applications for vision-based problems.
In contrast to traditional computer vision techniques with handcrafted features, deep neural networks can learn features directly from training data as a data-driven approach, achieving superior results than previous approaches.

\subsection{Pedestrian Detection}
Recent works on convolutional and recurrent neural networks enable deformation and occlusion handing \autocite{ouyang2013joint}, using Convolutional Neural Network (CNN) to detect objects of interest from UAVs imagery \autocite{bejiga2017convolutional}, and using multiple part detectors to combine as a pedestrian detector \autocite{tian2015deep}.
Object detections mainly utilise a sliding window approach or object proposal mechanism~\autocite{ren2015faster}.
Deep learning had tremendous successes in suppressing traditional feature-extraction methods in many other fields, for example, in
natural science~\autocite{xu2022waterSedimentML}, medical science~\autocite{sensorsMLforDiabetes}, or even in financial sectors~\autocite{forexNonStationaryTimeSeries}.
These methods demonstrated the representation power of deep learning on complicated problems, which are often hard to accomplish with handcrafted features.
However, they do not jointly reason the detected object with their temporal information, relying only on convolutional layers.
Recent works have framed object detection as a regression problem \autocite{redmon2016yolo} by densely generating bounding boxes with spatial separation afterwiards.

Robust human detection is an essential part of the UAV system as it facilitates capturing the locations of humans and reasoning on their current actions.
UAV imagery in object detections~\autocite{wang2018fast,zhang2020object,kraft2021autonomous} had been a specific subfield in computer vision.
Video feeds captured on UAVs are usually overhead shots with a wide variety of orientations, often highly varying compared to typical person detection~\autocite{hager2016combining,dike2018unmanned,sambolek2020person}.

\subsection{Activities Understanding}
Action understanding helps provide context on the detected person's current activities.
It is also a crucial aspect of a disaster response system as it provides contextual and situational information for the captured scenes~\autocite{yang2019effective}.
Most current approaches to multi-pedestrian action recognition take a sequential approach by separating each component and optimising each part separately.
Human is first detected in a CNN model and tracked with an algorithm, then feature representations are extracted for each person to reason on their action \autocite{ramanathan2016detecting}.
This approach requires much processing time as it needs to repeat the process for each detected person, which does not scale well on a UAV system.
A 3D CNN has also been proposed for action recognition \autocite{ji20133d}, which is based on a similar concept to regular CNNs with extra convolutions in the temporal axis.
Such an approach can often achieve high accuracy with the expenses of computational overhead~\autocite{sozykin2018multi,xu2017r}, often a limiting factor in UAV systems.
It achieves high accuracy in action understanding with the cost of substantial computational time.
Therefore, the limitation of computational power and batteries on modern UAVs restrict the technique's applicability to perform inferences directly onboard.

\subsection{Attentive Mechanism}
Recent works on the visual attention model gained a lot of empirical achievements, which is particularly important for interpreting the recognition of objects in cluttered scenes~\autocite{borji2012state}.
They act as a selection mechanism that encodes the notion of relevance regarding the overall rich stream of visual data.
Visual attention can guide the computational models, such as a neural network, to essential parts of the scene to gather more local information~\autocite{zheng2017learning}.
Attention map has been proven to be an effective way to allow neural nets to focus on local details that are more semantically important for enhancing overall results.
For example,~\autocite{hong2016learning} uses semantic segmentation as an attention map to transfer knowledge across the domain.
A context-attention map generated with another neural net can be utilised to improve human pose estimation~\autocite{chu2017multi}.
Gaussian Mixture Attention Maps have also previously been used as a depth feature to learn and reason jointly on multiple channels jointly~\autocite{wang2016differential}.
However, most approaches require a separate neural network to generate an additional attention layer, which is computationally prohibited on embedded systems with limited hardware.

We propose a simple yet effective way to produce a pseudo-attention map for our bounding box expansion.
Empirically, we show that expanding the bounding box helps improve the accuracy of temporal reasoning. Furthermore, the added attention map layer further improves the result by encoding the semantic spatial information to the expanded boxes.

\subsection{Embedded Systems in UAVs}
Most meaningful tasks for autonomous operations of UAVs require a combination of advanced sensors, complex image processing procedures and flight control algorithms~\autocite{al2018survey}.
Traditional algorithms and models that perform well in a ground station might not be suitable in the embedded system in UAVs.
For example, \autocite{hulens2015choose} discussed the trade-off between speed, power consumption and the weight of specific hardware platforms when operating in UAVs.
Advanced and computationally expensive algorithms are not directly applicable to operation on embedded devices.
Autonomous operations of drones require substantial computational resources on tasks such as SLAM~\autocite{lai2022slamreview} for localisation, motion planning~\autocite{lai2021rrf} of aircraft for planning trajectories, often in the kinodynamic space~\autocite{lai2021kinoSBP} for acceleration controls.
Therefore, using minimal computational resources on computer vision models is essential for the real-time operation of autonomous UAVs.
However, this approach uses numerical and graphical indicators to assess the vegetation properties instead of a deep-learning approach to maintain low computational overhead.
Successful usage of vision-based approaches in UAVs includes applzications such as Friesian cattle recovery~\autocite{andrew2019aerial}, counting vehicles in traffic~\autocite{amato2019counting}, and drone safe landing~\autocite{kakaletsis2021computer}.
The computational power and sensor requirements for UAVs' computer vision tasks are often different than that of an offline counterpart~\autocite{douklias2022design}.
Therefore, it is necessary to consider the physical hardware limitations when designing UAVs' computer vision applications.

\begin{figure}[bt]
    \centering
    \begin{tikzpicture}
        \node (img) {\includegraphics[width=.65\linewidth]{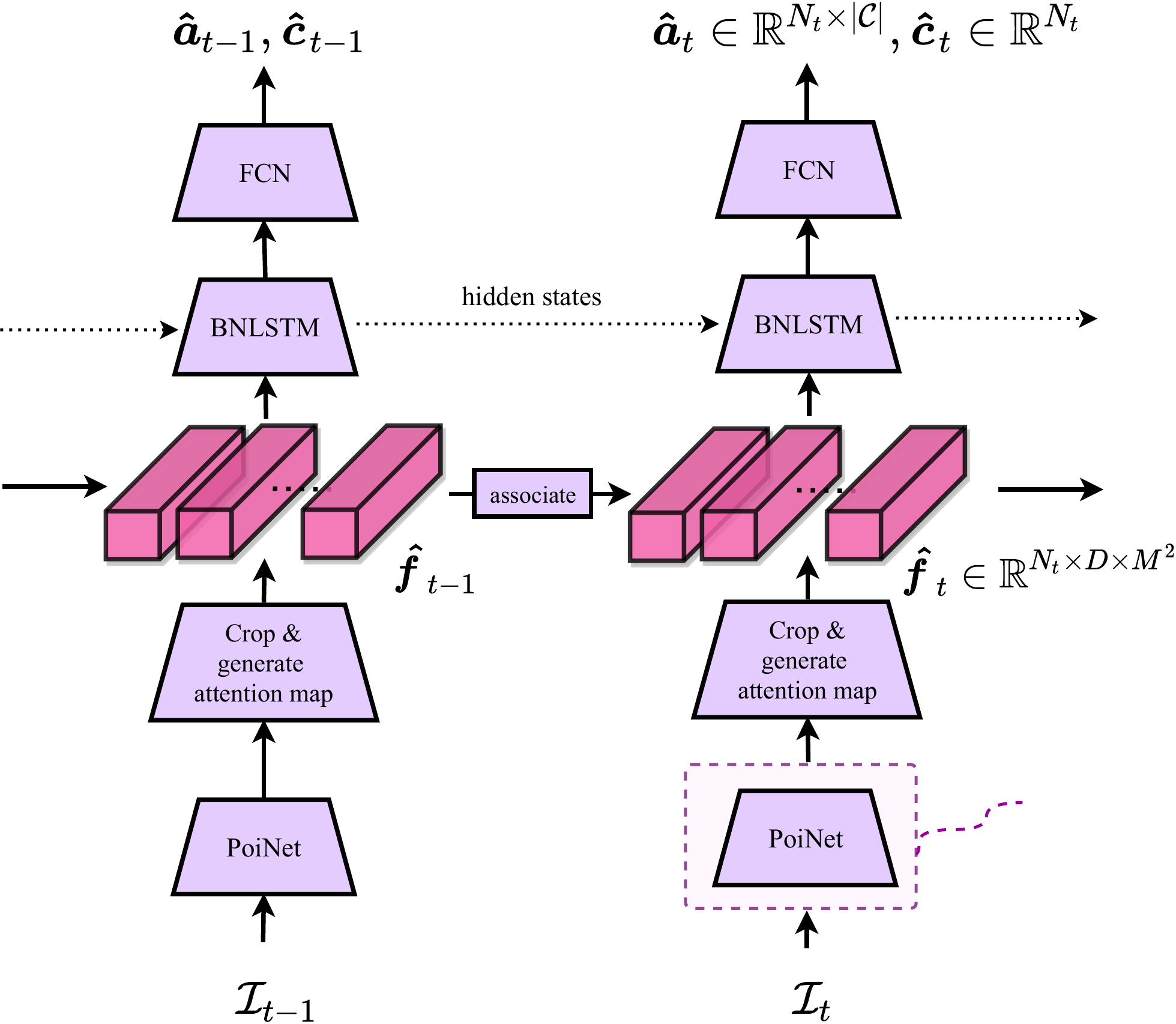}};
        \node at (4.50,-2.65){\Cref{fig:poinet}};
    \end{tikzpicture}%
    \caption{
        Overview of our entire framework.
        Frame $\mathcal{I}_t$ are queried at each time-step $t$ and passed through our \poinet (\cref{fig:poinet}) to generate bounding boxes $\vect{b}_t$ on the given frame, which is then used to crop on the original dense feature map to extract fixed-size $\vect{f}_t$.
        The extracted dense features are first concatenated with the generated attention map, then passed to a BNLSTM to reason on the temporal information.
        The final outputs are the probability of each recognised activity $\vect{a}_t$ and confidence scores $\vect{c}_t$ to suppress low confidence bounding box.
            \label{fig:entire-arch}
        }
\end{figure}

\section{Methodology}\label{sec:method}
We propose a unified architecture for a real-time system deployed in UAVs for autonomous surveying.
To this end, we designed our model to be lightweight and memory-efficient for usage on an embedded system.
We propose a detection network specifically for detecting pedestrians captured in different orientations and combined with a temporal network for jointly recognising individual activities.

\subsection{Overview}\label{sec:method-overview}
The overview of our entire architecture is shown in \cref{fig:entire-arch},
and overall description is given in \cref{alg:overall}.
Our architecture is composed of a pedestrian detection model, specifically designed for aerial surveying, as the entry point of the architecture.
Each video frame at time $t$, which we denoted with $\mathcal{I}_t$, is input into the detection network (\cref{sec:poinet}).
The output is a dense multiscale feature $\mathcal{F}_t$, of which we will extract the bounding boxes into a fixed size $\vect{f}_t\in\mathbb{R}^{N_t\times D\times M^2}$ representation.
The following components further extract temporal information from the dense features, which will be referred to as \emph{ActivityNet} (\cref{sec:activity-net}).
Finally, the model outputs multiple bounding boxes on pedestrians detection with associated multidimensional predicted actions and confidence scores.

\subsection{POINet}\label{sec:poinet}

This section will formulate the pipeline and architecture of our Position and Orientation Invariant detection Network (\poinet) model.
This model is responsible for producing bounding boxes of any detected pedestrian in our framework.
The challenge lies in the UAVs' operating camera angles, which often vary significantly due to the problem setup's nature.
We incorporated traditional computer vision in the generation of bounding boxes.
The use of fast and robust computer vision techniques is combined with the extracted features from CNN for its representation power.
Therefore, our approach uses less computational resources and time compared to approaches that use dense proposals of bounding boxes~\autocite{redmon2016yolo}.
An example of typical UAV imagery is shown in~\cref{fig:box-anchor-example:output}, where the size of pedestrians (and bounding boxes) are incredibly tiny compared to the overall frame size.
Furthermore, in aerial surveying, most video feeds are recorded in many different possible orientations, which creates severe challenges to robust detection.
For example,~\cref{fig:attention-map:other-examples} illustrates several examples of pedestrians captured in bird's-eye shots, high-angle shots, and near side-view shots.
\poinet addresses these issues by first extracting multiscale features, then using segmentation and regression maps for anchoring the location of each detected pedestrian.

\subsubsection{Anchoring Bounding Boxes}

The extracted frame~$\mathcal{I}_t$ from the input video feed is first fed into some lightweight CNN to extract high dimensional features.
To this end, we used MobileNetV2 for feature extractions aimed at embedded hardware by design \autocite{sandler2018mobilenetv2}.
During aerial surveying, pedestrians are often captured at different scales with no fixed zoom level.
Therefore, we extract high-dimensional features at different scales and then further concatenate them as multi-scale dense features $\mathcal{F}_t$.
With the extracted $\mathcal{F}_t$, \poinet uses basic convolutional layers to generate segmentation map $\mathcal{S}\in\mathbb{R}^{|W \times H|}$ and regression maps $\mathcal{R}\in\mathbb{R}^{|2\times W \times H|}$.
Semantically, the segmentation map $\mathcal{S}$ represents the area of interest that the \poinet is trained on;
whereas the regression map $\mathcal{R}$ contains the spatial information of the bounding boxes.
Segmentation map $\mathcal{S}$ and regression maps $\mathcal{R}$ are each given a ground truth that consists of crucial spatial information to reconstruct bounding boxes \autocite{bagautdinov2017social}.

\begin{figure}[tb]
    \centering
    \includegraphics[width=\linewidth]{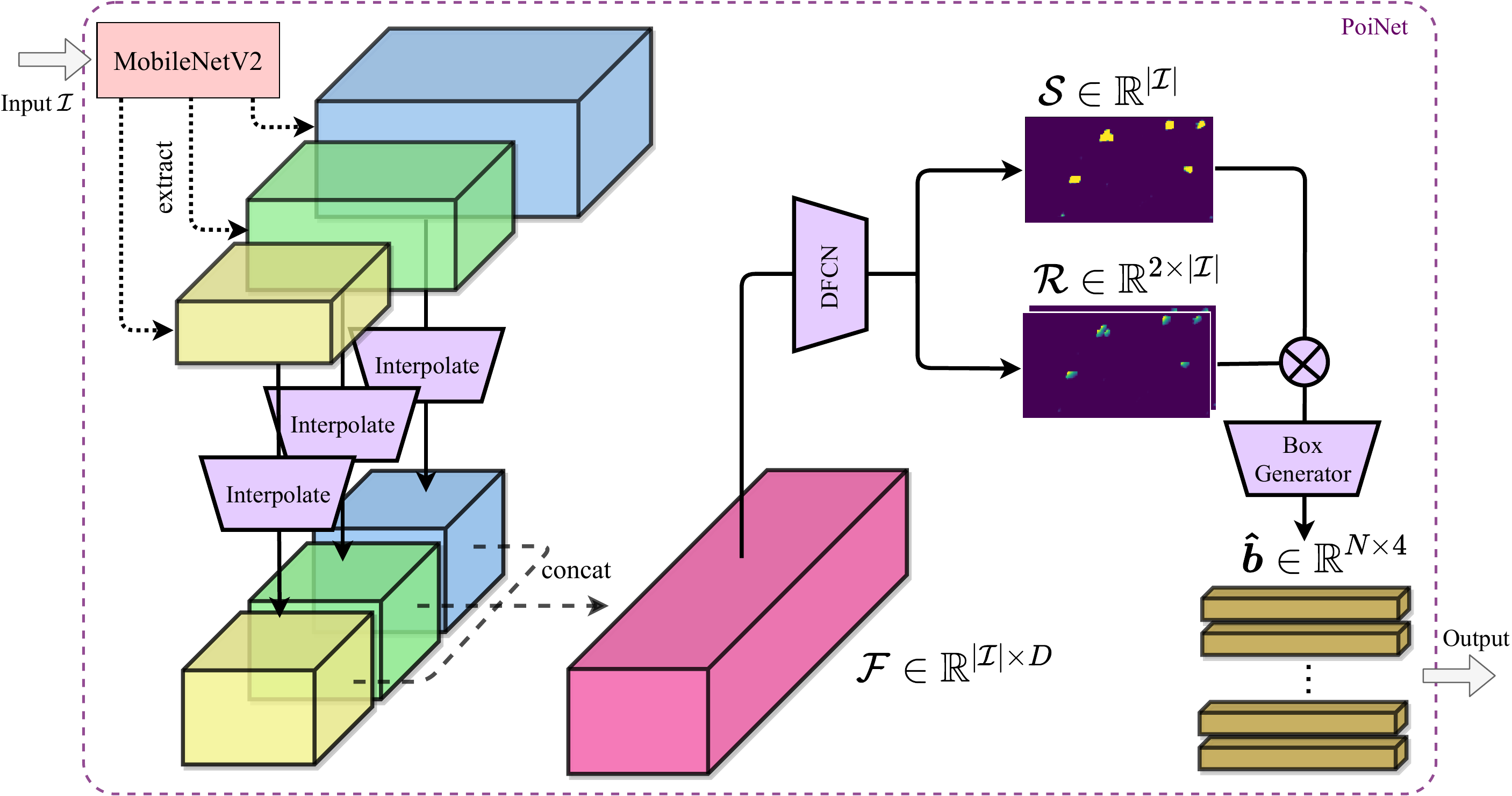}
    \caption{
    Overview of the \poinet architecture, detailing the transformations of the input image data for robust pedestrians detection.
        \label{fig:poinet}
    }
\end{figure}

\newcommand{\irow}[1]{%
  \begin{smallmatrix}[\,#1\,]\end{smallmatrix}%
}

We convert the ground truth pedestrian location into dense maps $\mathcal{S}$ and $\mathcal{R}$ with the following.
For each captured image $\mathcal{I}_t$ at time $t$, the ground truth consists of a set of bounding boxes $\vect{b}\in\mathbb{R}^{N_t\times4}$ where $b_i\in\vect{b}$ consists of the location information $(x_0,y_0,x_1,y_1)_t$ and $N_t$ is the number of groundtruth boxes for $\mathcal{I}_t$.
The $\mathcal{S}$ and $\mathcal{R}$ are defined over all the specific location $\vectalt{i}=(i_x,i_y)\in\mathcal{I}$, where we will use $\mathcal{S}_i,\mathcal{R}_i$ to denote their values at $(i_x,i_y)$.
For the segmentation map $\mathcal{S}$, we define %
\begin{equation}
 \mathcal{S}_i =
 \begin{cases}
    1 & \text{if } x_0 \le i_x \le x_1 \land y_0 \le i_y \le y_1 \\
    0 & \text{otherwise},
 \end{cases}
\end{equation}
such that it overlays areas with pedestrians.
\Cref{fig:box-anchor-example:seg} demonstrates an example of a segmentation map $\mathcal{S}$, which allows the model to learn the semantic patterns of pedestrians.
For the regression map $\mathcal{R}$, each location $(i_x,i_y)$ is a vector
$\mathcal{R}_i=\irow{r_0 & r_1}^\top$
that encodes the spatial location---top-left and bottom-right diagonal corners---of the bounding box, respectively.
We define $\mathcal{R}_i=\irow{0 & 0}^\top$ if $\mathcal{S}_i=0$.
For pixels $i$ within the segmented area (i.e. $\mathcal{S}_i=1$), we define $\mathcal{R}_i$ to be in the range of 0 to 1, such that
\begin{equation}
\mathcal{R}_i =
\begin{bmatrix}
    r_0 \\
    r_1
\end{bmatrix}
=
\frac{1}{\alpha}
\begin{bmatrix}
    x_1 - i_x & y_1 - i_y \\
    i_x - x_0 & i_y - y_0
\end{bmatrix}
\begin{bmatrix}
    \cos \theta \\
    \sin \theta
\end{bmatrix}
\end{equation}
where
\begin{equation}
 \theta = \arctan (y_1-y_0) \, / \, (x_1-x_0)
\end{equation}
is the angle of the diagonal and
\begin{equation}
    \alpha = \sqrt{(y_1-y_0)^2 + (x_1-x_0)^2}
\end{equation}
is the normalising constant.
\Cref{fig:box-anchor-example:reg1,fig:box-anchor-example:reg2} illustrate an example of the regression maps $\mathcal{R}$ which allow the model to spatially learn the anchor points for the bounding boxes of pedestrians.

During inference, we begin by first extracting the area of interest by obtaining $\mathcal{R'}$ in~\cref{alg:box-generator}~\cref{alg:bounding-box-area-of-interest}.
We will then remove noises in $\mathcal{R'}$ that only contain a tiny contiguous area of patches.
Afterwards, we apply a maximum filter across $\mathcal{R'}$ (\cref{fig:box-anchor-example:reg1,fig:box-anchor-example:reg2}) to obtain box corner coordinates, which are then used to densely generate multiple bounding boxes $\vect{b}$.
Finally, we programmatically remove boxes that contain less than $\delta\in\mathbb{R}, 0 < \delta \le 1$ fraction amount of segmented pixels (\crefrange{alg:bounding-box-fraction:start}{alg:bounding-box-fraction:end}), and practically we set $\delta=0.9$.

\begin{algorithm}[tb]
    \caption{Bounding box generator} \label{alg:box-generator}
    \KwPara{$\delta$: minimum percentage of segmented area}
    \Fn{$\FuncSty{BoxGenerator}(\mathcal{S}, \mathcal{R})$}{
        $\mathcal{R'} \gets \mathcal{S}\cdot\mathcal{R}$ \Comment*[r]{extracts area of interest\label{alg:bounding-box-area-of-interest}}
        remove noise (tiny area of patches) in $\mathcal{R'}$ \;
        $\vect{p} \gets$ find local maxima in $\mathcal{R'}$ via maximum filter\;
        \Comment{now $\vect{p_1}$ contains the top left and $\vect{p_2}$ contains the bottom right coordinates of the candidate boxes}
        $\vect{b} \gets$ combinations of $\vect{p_1}$ and $\vect{p_2}$ %
        \;
        \Comment{$\vect{b}$ is a set of box corner coordinates}
        $\vect{b'} \gets \emptyset$\;
        \ForEach{$b \in \vect{b}$}{
            $B \gets$ set of pixels enclosed by the box $b$\label{alg:bounding-box-fraction:start}\;
            $B' \gets \left\lbrace\;
            \vectalt{i} \in B \;\middle|\;
                 \begin{tabular}{@{}l@{}}
                    $\vectalt{i}$ is segmented as pixels \\
                    of interest in $\mathcal{S}$
                \end{tabular}
            \;\right\rbrace$\;
            \If{$|B'|~/~|B| \ge \delta$}{
                $\vect{b'} \gets \{b\} \cup \vect{b'}$\label{alg:bounding-box-fraction:end}\;
            }
        }
        \Return{$\vect{b'}$} \Comment*[r]{set of bounding boxes}
    }
\end{algorithm}

\subsubsection{Incorporating Temporal Information
 in Pedestrian Detection
 }

While spatial information is crucial for pedestrian detection, temporary changes can often negatively influence the detection accuracy due to changes in perspective or occlusion \autocite{wang2009hog}.
Therefore, \poinet is designed to densely generate bounding boxes and then jointly refined by the leaked temporal information from the activity layers.
\emph{ActivityNet} (\cref{sec:activity-net}) is located at the end of our pipeline (\cref{fig:entire-arch}), which uses temporal information to predict the current activity of each detected pedestrian and associated confidence scores $\vect{c}_t\in\mathbb{R}^{N_t}$.
The confidence scores are then used to jointly refine the bounding boxes $\vect{b}\in\mathbb{R}^{N_t\times4}$ with non-maximum suppression~\autocite{neubeck2006efficient}.
Non-maximum suppression is a technique to select one entity out of many overlapping entities based on the boxes' probability and the measure of the overlapping region with the Intersection over Union (IoU) metric. We can choose the selection criteria to arrive at the desired results. The leaked confidence scores from the activity component are used to help jointly enhance the refinement of the bounding box proposer.

\subsection{ActivityNet}\label{sec:activity-net}

The activity detection component of our \emph{ActivitytNet} framework takes input directly from \poinet to learn temporal information.
With each detected pedestrian, a novel attention map-based approach is first used to improve the effectiveness of learning activities (\cref{sec:attention-map}).
Afterwards, we pass the cropped dense pedestrian features to a Batch Normalised Long Short-Term Memory (BNLSTM) block~\autocite{bnlstm} as detailed in \cref{sec:train-end-to-end}.

\begin{figure}[!tb]
    \centering
    \begin{subfigure}{\linewidth}
        \centering
        \begin{subfigure}{.33\linewidth}
            \includegraphics[width=.99\linewidth]{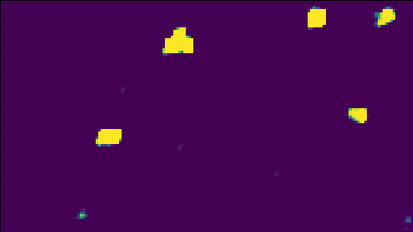}
            \caption{ Segmentation map \label{fig:box-anchor-example:seg}}
        \end{subfigure}%
        \begin{subfigure}{.33\linewidth}
            \includegraphics[width=.99\linewidth]{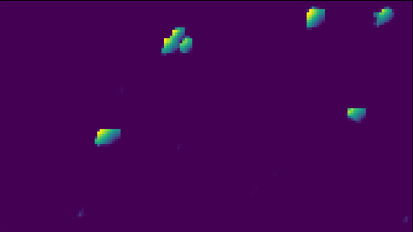}
            \caption{ Regression map 1 \label{fig:box-anchor-example:reg1}}
        \end{subfigure}%
        \begin{subfigure}{.33\linewidth}
            \includegraphics[width=.99\linewidth]{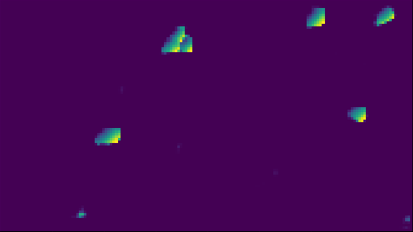}
            \caption{ Regression map 2 \label{fig:box-anchor-example:reg2}}
        \end{subfigure}
        \par\vspace{0.8mm}
        \begin{subfigure}{.5\linewidth}
            \includegraphics[width=.99\linewidth]{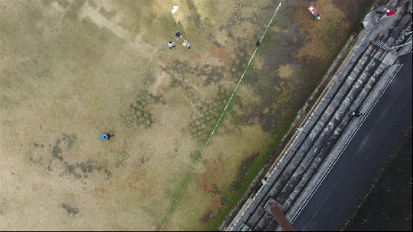}
            \caption{Input video frame \label{fig:box-anchor-example:input}}
        \end{subfigure}%
        \begin{subfigure}{.5\linewidth}
            \includegraphics[width=.99\linewidth]{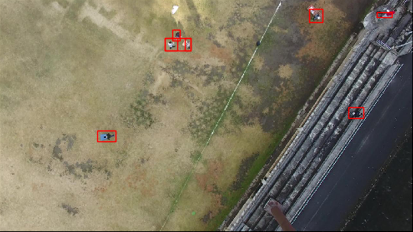}
            \caption{Output bounding boxes \label{fig:box-anchor-example:output}}
        \end{subfigure}
    \end{subfigure}
    \caption{
        Generating bounding boxes by extracting spatial information from the images.
        (\subref{fig:box-anchor-example:input}) Captured video frames are first feed into \poinet.
        The network will outputs (\subref{fig:box-anchor-example:seg}) segmentation map and (\subref{fig:box-anchor-example:reg1}, \subref{fig:box-anchor-example:reg2}) regression maps of which summarised the points of interest that the \poinet is trained on.
        The maps will then be passed to several filters and suppression algorithms to output (\subref{fig:box-anchor-example:output}) final bounding boxes.
        \label{fig:box-anchor-example}
    }
\end{figure}

\subsubsection{Attention map}\label{sec:attention-map}

We employ a novel attention map-based technique~\autocite{zheng2017learning} to facilitate adding additional contextual information to each bounding box.
It enables a programmatic way to incorporate the bounding boxes' surrounding environment while maintaining the spatial information of the boxes.
With the approach in \autocite{bagautdinov2017social} of directly using the cropped boxes as the input features, the boxes will lose much contextual information in the background---which is arguably essential for learning the activities.
Here we are employing a simple yet effective approach to incorporate more contextual information and the bounding box.
In addition, the attention map approach also allows us to avoid introducing distortion of aspect ratio for the cropped images, which is common in many RNN approaches.
However, previous studies found that distorted aspect ratio region tends to give higher false-positive bounding boxes~\autocite{xu2017deformable}, where preserving the shape of objects is highly beneficial for accurate classification~\autocite{zheng2016good}.

The attention matrix $A$ uses a flexible multivariate Gaussian kernel to encode the likely correlation of the expanded region of the bounding box while allowing our temporal network to reason on previously absent contextual information.
We define the attention matrix as $A \in \mathbb{R}^{M \times M}$ where
$
    M = \lceil \alpha \cdot \max(W,H) \rceil
$
 is the dimension of the expanded bounding box, and $\alpha\in\mathbb{R},\alpha \ge 1$ is a factor that controls the ratio of expansion.
The values of the attention matrix are defined as
\begin{equation} \label{eq:attention-map-val}
 A_i =
 \begin{cases}
    1 & \text{if } x_0 \le i_x \le x_1 \land y_0 \le i_y \le y_1 \\
    \varphi(\vectalt{i} \,|\, \vectalt{b_c}, \vectalt{\Sigma} ) & \text{otherwise},
 \end{cases}
\end{equation}
where
    $
    \varphi(\vectalt{x} \,|\, \vectalt{\mu},\vectalt{\Sigma}) =
    \exp\left\{ -\frac{1}{2} (\vectalt{x} - \vectalt{\mu}) \vectalt{\Sigma}^{-1} (\vectalt{x} - \vectalt{\mu}) \right\}
    $
resembles the probability density distribution of the multivariate Gaussian~\autocite{bishop2006pattern} without normalising factor,
Moreover, $\vectalt{\Sigma}$ is a positive definite covariance matrix that encodes our belief on the spatial correlation between the action of the pedestrians and their surrounding environment.
\Cref{sec:experimental-result} provides experimental results on the effectiveness of the additional attention map.
The flexibility of this novel attention map allows us to express the relevance of each pixel to the surrounding contextual information.
We formulate an algorithmic way to generate a pseudo-attention map based purely on the given bounding box ground truth.
In addition, this novel method directly avoids distortion of cropped images, and experimentally, we show that the additions of the attention map enhance the overall model accuracy.

\begin{figure}[!tb]
    \centering
    \begin{subfigure}{.75\linewidth}
        \centering
        \begin{subfigure}{.4\linewidth}
            \includegraphics[width=.99\linewidth]{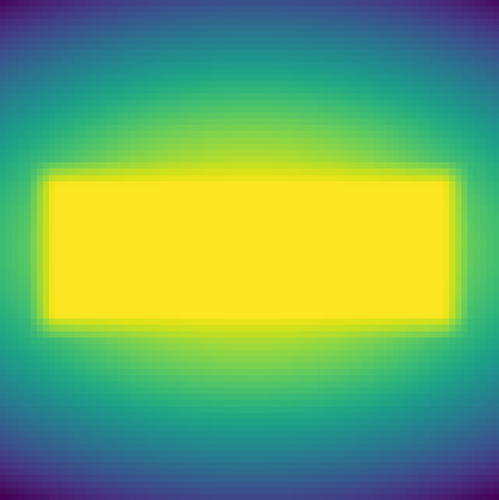}
            \caption{Example of an attention map \label{fig:attention-map:example}}
        \end{subfigure}\hspace{2mm}%
        \begin{subfigure}{.062\linewidth}
            \includegraphics[width=.99\linewidth]{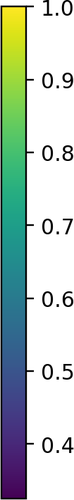}%
            \vspace{5mm}
        \end{subfigure}\hspace{7mm}%
        \begin{subfigure}{.4\linewidth}
            \includegraphics[width=.99\linewidth]{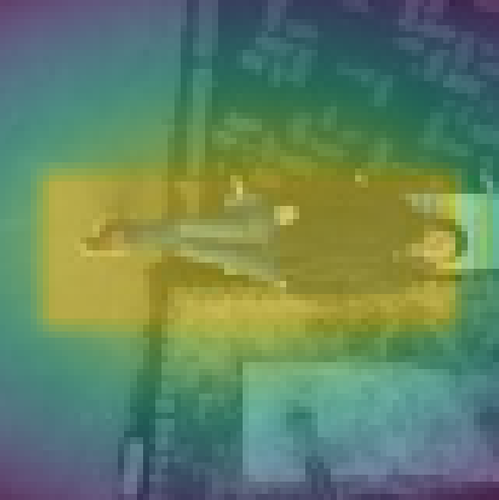}
            \caption{Corresponding pedestrian of (\subref{fig:attention-map:example})  \label{fig:attention-map:exmaple-overlay}}
        \end{subfigure}
    \end{subfigure}

    \par\vspace{0.8mm}

    \begin{subfigure}{.85\linewidth}
        \includegraphics[width=.33\linewidth]{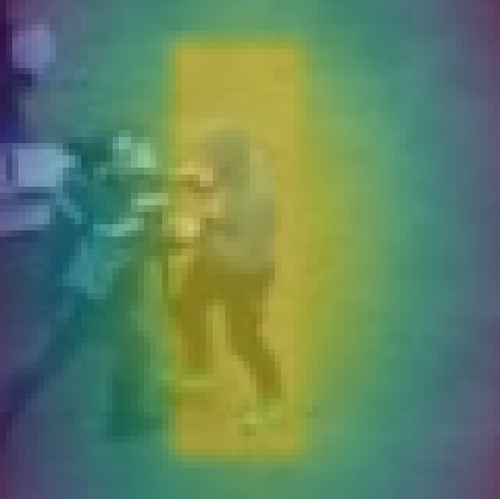}\hfill%
        \includegraphics[width=.33\linewidth]{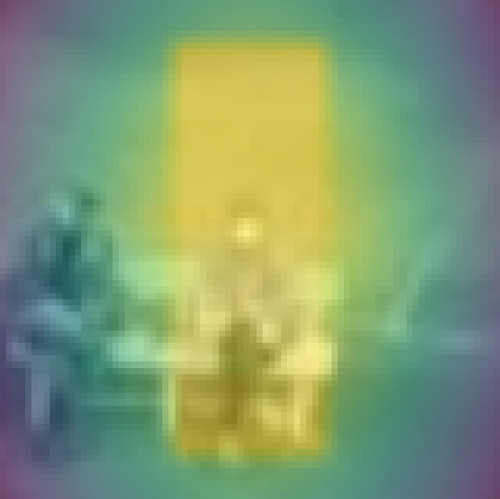}\hfill%
        \includegraphics[width=.33\linewidth]{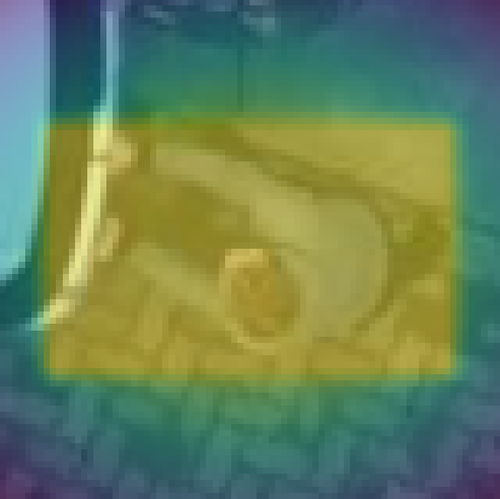}\hfill%
    \par\vspace{0.4mm}
        \includegraphics[width=.33\linewidth]{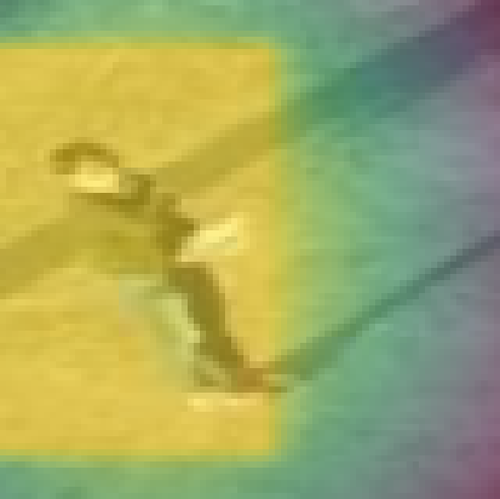}\hfill%
        \includegraphics[width=.33\linewidth]{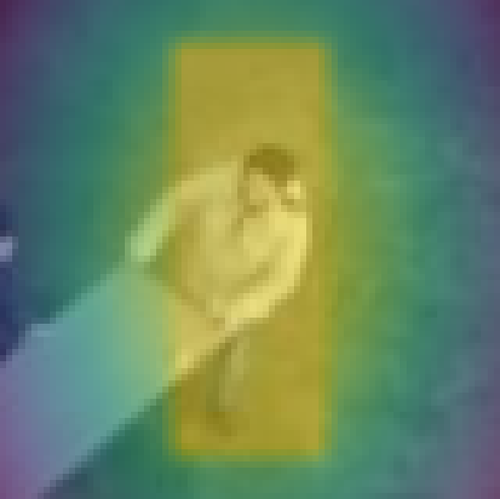}\hfill%
        \includegraphics[width=.33\linewidth]{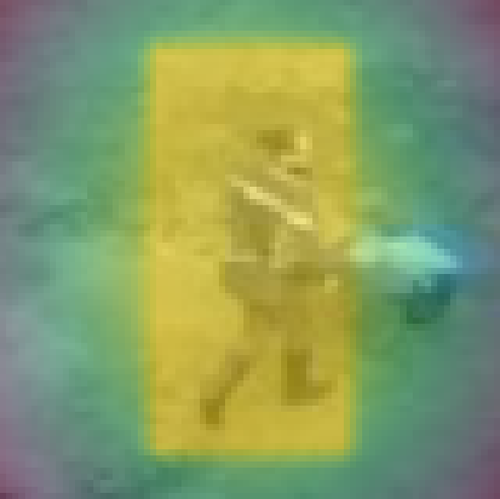}\hfill%
        \caption{Other examples of cropped bounding boxes with attention map overlay \label{fig:attention-map:other-examples}}
    \end{subfigure}

    \par\vspace{1.5mm}

    \caption{
        Attention maps are generated to denote which portion of pixels is more significant for the widened bounding boxes.
        The yellow rectangular shape in each example can be regarded as the original bounding box.
        A standalone example of an attention map is shown in (\subref{fig:attention-map:example}), with the corresponding cropped image shown in (\subref{fig:attention-map:exmaple-overlay}).
        More examples of attention maps are shown in (\subref{fig:attention-map:other-examples}), which showcase instances that a widened bounding box will add valuable contextual information to the neural networks.
        \label{fig:attention-map}
    }
\end{figure}

\subsubsection{Training \poinet and ActivityNet End-to-End}\label{sec:train-end-to-end}
We have described a way to obtain a set of reliable detection from raw images from previous sections.
However, temporal information was shown to be important in action recognition and provide a valuable basis for reasoning on detection.
Therefore, after obtaining predictions by the BNLSTM component in~\cref{fig:entire-arch} we utilise the temporal confidence score $\vect{c}_t$ by jointly refining the detected bounding boxes.
That is, we use confidence scores to refine and suppress low confidence bounding box with non-maximum suppression~\autocite{neubeck2006efficient}.
The reason is that temporal information can often provide a better estimate of whether the captured bounding box contains a pedestrian.

For each frame $\mathcal{I}_t$, given that we had extracted $N$ bounding boxes of pedestrian $\vect{b}_t = \{\vectalt{b}_t^1,\ldots,\vectalt{b}_t^N\}$, we extract a fixed-sized representation $\vect{f}_t = \{f_t^1,\ldots,f_t^N\}$ for each box from the multiscale dense feature $\mathcal{F}_t$.
These embeddings are then sent to our BNLSTM block, which produces predictions on their corresponding activities $\vect{a}_t$ and confidence scores $\vect{c}_t$ that denote the likelihood that the sequence of frames contains actual pedestrians.
Finally, we construct a loss that accounts for the multi-activities nature of the dataset.
Each pedestrian can be executing a \emph{primary} and a \emph{secondary} action simultaneously.
For example, a pedestrian can be \emph{walking} while \emph{pushing} something, or \emph{sitting} while \emph{reading} a book.
Let $\vectbar{p}^i_{t,p}$, $\vectbar{p}^i_{t,s}$ for $i\in\{1,\ldots,N_t\}$ be the corresponding one-hot-encoded ground truth of \emph{primary} and \emph{secondary} activities, where $N_t$ is the number of bounding box at frame $t$.
We will match our predictions using the closest bounding box distance, which gives us the prediction vector $\vectalt{p}_{t,p}^i$ and  $\vectalt{p}_{t,s}^i$ for $i\in\{1,\ldots,N_t\}$.
We can now express the loss function as
\begin{equation}
    \mathcal{L}_{ps} =
    -
    \frac{1}{T} \sum^T_{t=1}  \Big(
        \frac{1}{N_t \cdot N_p} \sum_{i=1}^{N_t} \vectbar{p}_{t,p}^i \log \vectalt{p}_{t,p}^i
        -
        \lambda_w
        \frac{1}{N_t \cdot N_s} \sum_{i=1}^{N_t} \vectbar{p}_{t,s}^i \log \vectalt{p}_{t,s}^i
        \Big)
\end{equation}
where $T$ is the total number of frames, $N_p$, $N_s$ are the number of labels for primary and secondary actions, and $\lambda_w$ is a weight factor that allows us to balance the two activities differently.
In practice, we set $\lambda_w=0.5$ to focus more on the \emph{primary} action.

\section{Experimental Results}\label{sec:experimental-result}

In this section, numerical results are reported on pedestrian detection and multi-action recognition tasks, comparing multiple baselines.
We also provide empirical results on the deployment of our networks on an embedded device.
In particular, the pipeline architecture is deployed on live UAVs to test its field performance and transmission speed.
The performance and practicality of the system are evaluated in this section.
We used an IoU of 0.5 as the threshold for the following results.

\subsection{Dataset}
We evaluate our architecture on the \emph{Okutama-Action} dataset \autocite{barekatain2017okutama}.
It is
a
publicly available multi-pedestrian drone dataset that contains labels at each frame that describe the pedestrian's current activities.
Deploying a reliable drone-based response system requires recognising the pedestrians' active state to detect their distress levels.
Other aerial surveying datasets exist but do not contain activity labels.
The dataset is a high-resolution video dataset captured by UAVs with varying camera angles and altitudes.
The dataset consists of 77K fully-annotated frames, where the video sequences are in 4K resolution and 30FPS.
Multiple pedestrians exist in a given frame, with labels describing the pedestrians' current activities.
The frames contain varying altitudes, ranging from about 10 to 45 meters, and varying camera angles from 45 degrees to 90 degrees (bird eyes view).
The current dataset consists of a labelled bounding box around each person and one or more person's actions. For example, a person can perform one \textit{primary} action, such as standing or walking, and one \textit{secondary} action, such as carrying or pulling.
\subsection{Baselines Models}
The following baselines are considered for comparing model performance in our drone dataset.

\textbf{Models for \textit{multi-activities prediction}} we evaluate against
(i) SSD RGB~\autocite{liu2016ssd}: pre-trained on ImageNet and fine-tuned to predict individual actions based on fixed-sized images of individual pedestrians;
(ii) SSD Optical Flow~\autocite{barekatain2017okutama}: a variant of the previous baseline that uses motion cues from the optical flow steam;
(iii) SSD-MobileNetV2~\autocite{sandler2018mobilenetv2}: similar to SSD RGB baseline, but instead MobileNet is used as the feature extractor as a light-weight variant;
(iv) Inception-SSD~\autocite{chen2020inception}: similar to the previous baseline, but instead Inception block is added as an extra layer in the SSD before the prediction;
(v) YOLOv3~\autocite{redmon2018yolov3}: pre-trained and fine-tuned to the target drone dataset, with an additional LSTM block at the end;
(vi) RAER~\autocite{ibrahim2018hierarchical}: a hierarchical autoencoder-based model that combines scene information with individual action. We follow the main proposed architecture of 4 relational layers and focus on the whole scene.

\textbf{Models for \textit{pedestrian detection}}
we evaluate against
(i) ReInspect~\autocite{stewart2016end}: a joint multi-person detection designed for crowded scene;
(ii) ReInspect-rezoom~\autocite{stewart2016end}: similar to the last baseline, but with an additional re-zooming layer that transforms features into a scale-invariant representation;
(iii) Faster-RCNN~\autocite{ren2015faster}: a widely adopted region proposal model;
(iv) SSD~\autocite{szegedy2016rethinking}: similar to the model that we evaluate in activities prediction, but focus only on proposing bounding boxes;
(v) YOLOv3~\autocite{liu2016ssd}: similarly focuses on bounding boxes proposal.

\begin{table}[!bt]
\begin{center}
{
{%
\begin{tabular}{@{}r@{}lcc@{}}
\toprule
\multicolumn{2}{c}{Method} & \begin{tabular}[c]{@{}c@{}}mAP (\%) \\ (\emph{primary} / \emph{secondary})\end{tabular}  & Inf. time (sec) \\ \midrule
{Ours} & \multicolumn{1}{l|}{}                                 & 24.2 / 13.6  & $1.43 \pm 0.21 $                              \\
{Ours} & \multicolumn{1}{l|}{(w/o att.)}                                 & 22.5 / 12.3  & $1.41 \pm 0.23 $                              \\
{\smaller SSD RGB}     & \multicolumn{1}{l|}{\autocite{liu2016ssd}}       & 18.8 / 7.41 & $2.02 \pm 0.39$                                 \\
{\smaller SSD Optical Flow}     & \multicolumn{1}{l|}{\autocite{liu2016ssd}}       & 6.47 / 2.21 & $1.91 \pm 0.37$                                 \\
{\smaller SSD-MobileNetV2}     & \multicolumn{1}{l|}{\autocite{sandler2018mobilenetv2}}       & 17.2 / 7.10 & $1.88 \pm 0.22$                                 \\
{\smaller Inception-SSD}     & \multicolumn{1}{l|}{\autocite{chen2020inception}}       & 19.6 / 9.51 & $2.02 \pm 0.39$                                 \\
{\smaller YOLO}v3-LSTM  & \multicolumn{1}{l|}{\autocite{redmon2018yolov3}} & 20.2 / 9.96  & $2.21 \pm 0.65 $                          \\
{\smaller RAER}  & \multicolumn{1}{l|}{\autocite{ibrahim2018hierarchical}} & 21.5 / 10.56  & $3.16 \pm 0.64 $                          \\ \bottomrule
\end{tabular}
 }
}
\end{center}
\caption{Multi action recognition performed on the embedded system TX2.
Inference time is in ($\mu \pm 2\sigma$) seconds.
\label{table:action-recognition}}
\end{table}

\subsection{Implementation Details}
All of our models are trained using the same optimisation and backpropagation scheme.
We use the Adam stochastic gradient descent optimiser, with an initial learning rate set to $10^{-4}$ and fixed hyperparameters to $\beta_1=0.9$, $\beta_2=0.999$, $\epsilon=10^{-8}$.
We begin by training the pedestrian detection component of each network, which involves using the given ground truth box coordinates to propose bounding boxes.
After a fixed number of epochs, we jointly train the temporal components.
Gradients are allowed to pass through the previous part, which facilitates each component to compensate for the other in an end-to-end setting.

\subsection{Model Deployment on Embedded-UAV}

\begin{figure}[tb]
    \centering
    \includegraphics[width=.75\linewidth]{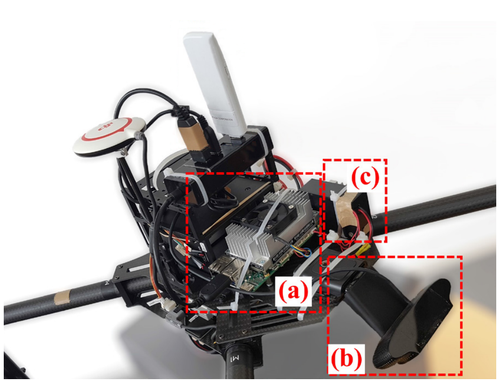}
    \caption{
    DJI Matrice 100 drone with an embedded TX2 device used for the field test and inference benchmark.
    (a) NVIDIA Jetson TX2, (b) Logitech C920 camera, (c) voltage regulator.
        \label{fig:drone-photo}
    }
\end{figure}

The overall drone and the onboard system are illustrated in~\cref{fig:drone-photo}.
The pipeline architecture is first trained on an external server to speed up training time.
The coding environment is developed on top of the Python Deep-Learning \emph{PyTorch} library.
Then, the trained model is converted to a Caffe model that can be run natively without the Python runtime environment.
We packaged the model as an onboard application and deployed it to the NVIDIA
Jetson TX2.
The NVIDIA Jetson TX2 is an embedded AI computing device that enables edge computing with an NVIDIA Pascal GPU with 8 GB RAM.
The Jetson TX2 then acted as the onboard computer and attached to the DJI Matrice 100 (\cref{fig:drone-photo}a).
The TX2 is mounted on the Auvidea J120 carrier board and connects
to the UAV using a USB to TTL/UART connection.
We attached a Logitech C920 camera (\cref{fig:drone-photo}b) to capture video footage, and an LTE dongle to the Jetson TX2 to communicate information back to the user.
The onboard battery for the UAV is a DJI TB48D battery with 130 watt-hours, acting as the power source to the onboard computer through a voltage regulator (\cref{fig:drone-photo}c) to maintain a constant voltage for the embedded board.
The USB camera capture footage and streams it to the Caffe model to perform inference.
The total payload of the entire onboard setup during the field test (\cref{fig:intro-image}) is at around 0.57 kg.

During deployment, an image frame is first fetched from the video feed, resized, and sent to the Caffe model.
The frame extraction procedure is performed using the OpenCV C++ library, typically taking $< 5$ milliseconds.
Then, the frame is passed to the \poinet, followed by the ActivityNet for the whole inference procedure.
The embedded processing device typically takes around 2-3 seconds to process each frame and transmit the corresponding information.
Depending on the number of detected pedestrians, the drone will then uses around $100-500$ bytes of network bandwidth to send the information back to the remote operator.

\subsection{Multi-Pedestrian Multi-activities Recognition}

In \cref{table:action-recognition}, we report the mean average precision (mAP) and inference time on the NVIDIA Jetson TX2 embedded on UAV during deployment.
We report accuracies separately for the primary and secondary actions by the pedestrians.
We take the top two predicted outputs as primary and secondary actions for models that do not predict multiple actions.
We also report a variant of our model without the additional attention map (\cref{sec:attention-map}) to evaluate its effectiveness.
Inference time performed on the embedded device reports the mean and the 95\% quantile over 30 runs in seconds.

Overall, our model achieves state-of-the-art performance for both \emph{primary} and \emph{secondary} actions recognition even without the ground truth locations of the individuals.
The comparison to SSD variants, which propose pedestrians' action based purely on captured frames, indicates that having a temporal joint representation shared across multiple tasks helps us improve accuracy.
YOLOv3-LSTM and RAER perhaps are the main competitor as they utilise temporal information across frames.
Of the two, RAER performs better as it utilises a hierarchical structure and scene information.
However, the architecture also imposes more overhead in inference time, which is essential in a real-time UAV application with limited processing power.
In fact, our model requires the lowest inference time as our model requires fewer parameters and uses operations that cut down runtime.
Our attention map approach also helps to improve performance, which is likely due to the extra information around the boundary that would otherwise be cropped (\cref{fig:attention-map}). Such an approach also helps to maintain the aspect ratio of the input to our temporal component.
Our algorithmic attention map will likely be less performant than a learning approach; however, it requires little to no overhead to generate, which is essential in time-sensitive applications.
\Cref{fig:intro-image} and \Cref{fig:box-anchor-example} are results from our actual running model.

\subsection{Multi-Pedestrian Detection}

In addition to evaluating the overall performance of action recognition, we also conducted a standalone experiment evaluating the performance for \poinet to propose bounding boxes from UAV imagery.
Our evaluation of pedestrian detection from the UAV dataset (\cref{table:bounding-box-detection}) found that existing box proposal approaches do not work well in sparse scenes with varying camera angles.
For example, our leading contender ReInspect works well in a dense proposal of a crowded scene as shown in~\autocite{stewart2016end}.
However, it does not perform well in UAV imagery, mainly when the objects of interest are sparsely spaced, which introduces more false positives in the model output.
The addition of an invariant layer in ReInspect-rezoom helps enhance the performance as it scales better with smaller objects.
However, both models require quite a significant runtime overhead due to the need to make joint predictions in a sequential manner.
It should be noted that SSD, YOLO and RCNN models are built to propose bounding boxes for various objects.
That is, they are capable of proposing bounding boxes with more than one class as they are designed for dense proposals of day-to-day objects.
However, for a domain-specific application where we are only interested in search and rescue in a sparse environment (e.g. in a wide open field), these model assumptions do not seem to perform well.
Therefore, using them solely for pedestrian detection seems to limit their predictability in our UAV application.
By exploiting the domain specificity, our proposed \poinet achieves higher precision with less computational overhead.

\begin{table}[!tb]
    \begin{center}
    {
    \begin{tabular}{@{}r@{}lcc@{}}
    \toprule
    \multicolumn{2}{c}{Method} & mAP (\%)  & Inf. time (sec) \\ \midrule
    \poinet & \multicolumn{1}{l|}{(ours)}                                 & 84.5  & $1.52 \pm 0.16 $                              \\
    {\smaller ReInspect}     & \multicolumn{1}{l|}{\autocite{stewart2016end}} & 79.8  & $3.41 \pm 0.77$                                 \\
    {\smaller ReInspect-rezoom}     & \multicolumn{1}{l|}{\autocite{stewart2016end}} & 81.1  & $3.50 \pm 0.69$                                 \\
    {\smaller Faster-RCNN}     & \multicolumn{1}{l|}{\autocite{ren2015faster}}       & 68.1  & $2.98 \pm 0.42$                                 \\
    {\smaller SSD}     & \multicolumn{1}{l|}{\autocite{liu2016ssd}}       & 72.3  & $1.98 \pm 0.37$                                 \\
    {\smaller YOLO}v3  & \multicolumn{1}{l|}{\autocite{redmon2018yolov3}} & 77.2  & $2.38 \pm 0.52 $                          \\
     \bottomrule
    \end{tabular}
}
    \end{center}
    \caption{Results for multi-pedestrian detection from UAVs imagery.
    Inference time is in ($\mu \pm 2\sigma$) seconds.
    \label{table:bounding-box-detection}}
\end{table}

\section{Conclusions}
We have proposed a novel approach for real-time UAV applications by exploiting domain knowledge.
Our model is designed to excel in detecting pedestrians regardless of their orientation from UAVs perspective.
It is lightweight and combines traditional computer vision techniques to acquire the anchor points of boxes: harnessing the rich representation power of the deep learning model for robust drone operations.

The constraints of performing inference directly onboard can severely limit the potential model complexity.
In our deployment, when our framework detects pedestrians and decides to report their status, it will transmit a message of 100–500 bytes to its operator (depending on the number of detected pedestrians).
Compared to streaming videos to a remote server for inference, it will take a continuous stream of images, each with about 200 kB.
Constructing a jointed detection and recognition model with constraints on output message size ensures model deployment's practicality in a distributed and reliable manner.

\subsection{Limitations}
Our model's limitations include only focusing on detecting pedestrians, as opposed to general objects like other approaches. Moreover, it is likely that our bounding box proposal mechanism would not perform well in cluttered scenarios, which is uncommon for disaster response in a remote area.
For future work, for a more UAV domain-specific approach, we can utilise altitude or velocity information from drone sensors as model features to provide more details during inference.
This approach might allow the model to account for the problematic scenarios and uncertainty issues in UAV imagery.

\subsection{Potential negative societal impact}
While the usage of UAVs in rural environments for disaster responses has been proven effective, the emergence of ubiquitous consumer UAV hardware has raised concerns. The learning of a pedestrian tracking and reasoning model in autonomous UAVs can have potential negative impacts on privacy issues, for example, deployments of aerial surveillance in cities.
Potential negative impacts can be minimised by imposing governmental policies and social-wise regulations.
In UAV-based anomaly detection, we can impose restrictions that the extracted pedestrians' information is only processed directly onboard.
If the UAV detects anomalies that require further actions by human operators, UAVs will only then transmit anonymised data to operators.

\printbibliography

@INPROCEEDINGS{abdelkader2013_UAVBase,
  AUTHOR = {Abdelkader, Mohamed and Shaqura, Mohammad and Claudel, Christian G. and Gueaieb, Wail},
  BOOKTITLE = {2013 International Conference on Unmanned Aircraft Systems ({{ICUAS}})},
  DATE = {2013-05},
  DOI = {10.1109/icuas.2013.6564670},
  EVENTTITLE = {2013 International Conference on Unmanned Aircraft Systems ({{ICUAS}})},
  TITLE = {A {{UAV}} Based System for Real Time Flash Flood Monitoring in Desert Environments Using {{Lagrangian}} Microsensors},
}

@INPROCEEDINGS{aljehani2016multi,
  AUTHOR = {Aljehani, Maher and Inoue, Masahiro},
  ORGANIZATION = {IEEE},
  BOOKTITLE = {2016 {{IEEE}} 5th {{Glob}}. {{Conf}}. {{Consum}}. {{Electron}}.},
  DATE = {2016},
  DOI = {10.1109/gcce.2016.7800524},
  TITLE = {Multi-{{UAV}} Tracking and Scanning Systems in {{M2M}} Communication for Disaster Response},
}

@INPROCEEDINGS{amato2019counting,
  AUTHOR = {Amato, Giuseppe and Ciampi, Luca and Falchi, Fabrizio and Gennaro, Claudio},
  ORGANIZATION = {IEEE},
  BOOKTITLE = {2019 {{IEEE Symp}}. {{Comput}}. {{Commun}}. {{ISCC}}},
  DATE = {2019},
  DOI = {10.1109/ISCC47284.2019.8969620},
  TITLE = {Counting Vehicles with Deep Learning in Onboard Uav Imagery},
}

@INPROCEEDINGS{andrew2019aerial,
  AUTHOR = {Andrew, William and Greatwood, Colin and Burghardt, Tilo},
  ORGANIZATION = {IEEE},
  BOOKTITLE = {2019 {{IEEERSJ Int}}. {{Conf}}. {{Intell}}. {{Robots Syst}}. {{IROS}}},
  DATE = {2019},
  DOI = {10.1109/IROS40897.2019.8968555},
  TITLE = {Aerial Animal Biometrics: {{Individual}} Friesian Cattle Recovery and Visual Identification via an Autonomous Uav with Onboard Deep Inference},
}

@INPROCEEDINGS{bagautdinov2017social,
  AUTHOR = {Bagautdinov, Timur and Alahi, Alexandre and Fleuret, François and Fua, Pascal and Savarese, Silvio},
  BOOKTITLE = {Proceedings of the {{IEEE}} Conference on Computer Vision and Pattern Recognition},
  DATE = {2017},
  DOI = {10.1109/CVPR.2017.365},
  TITLE = {Social Scene Understanding: {{End-to-end}} Multi-Person Action Localization and Collective Activity Recognition},
}

@INPROCEEDINGS{bai2010emergency,
  AUTHOR = {Bai, Yong and Du, Wencai and Ma, Zhengxin and Shen, Chong and Zhou, Youling and Chen, Baodan},
  ORGANIZATION = {IEEE},
  BOOKTITLE = {2010 {{IEEE Int}}. {{Conf}}. {{Wirel}}. {{Commun}}. {{Netw}}. {{Inf}}. {{Secur}}.},
  DATE = {2010},
  TITLE = {Emergency Communication System by Heterogeneous Wireless Networking},
}

@INPROCEEDINGS{barekatain2017okutama,
  AUTHOR = {Barekatain, Mohammadamin and Martí, Miquel and Shih, Hsueh-Fu and Murray, Samuel and Nakayama, Kotaro and Matsuo, Yutaka and Prendinger, Helmut},
  BOOKTITLE = {Proceedings of the {{IEEE}} Conference on Computer Vision and Pattern Recognition Workshops},
  DATE = {2017},
  DOI = {10.1109/CVPRW.2017.267},
  TITLE = {Okutama-Action: {{An}} Aerial View Video Dataset for Concurrent Human Action Detection},
}

@ARTICLE{bejiga2017convolutional,
  AUTHOR = {Bejiga, Mesay and Zeggada, Abdallah and Nouffidj, Abdelhamid and Melgani, Farid},
  PUBLISHER = {Multidisciplinary Digital Publishing Institute},
  DATE = {2017},
  DOI = {10.3390/rs9020100},
  JOURNALTITLE = {Remote Sens.},
  NUMBER = {2},
  PAGES = {100},
  TITLE = {A Convolutional Neural Network Approach for Assisting Avalanche Search and Rescue Operations with Uav Imagery},
  VOLUME = {9},
}

@BOOK{bishop2006pattern,
  AUTHOR = {Bishop, Christopher M},
  PUBLISHER = {springer},
  DATE = {2006},
  TITLE = {Pattern Recognition and Machine Learning},
}

@ARTICLE{borji2012state,
  AUTHOR = {Borji, Ali and Itti, Laurent},
  PUBLISHER = {IEEE},
  DATE = {2012},
  DOI = {10.1109/tpami.2012.89},
  JOURNALTITLE = {IEEE Transactions on Pattern Analysis and Machine Intelligence},
  NUMBER = {1},
  PAGES = {185--207},
  SHORTJOURNAL = {IEEE Trans. Pattern Anal. Mach. Intell.},
  TITLE = {State-of-the-Art in Visual Attention Modeling},
  VOLUME = {35},
}

@ARTICLE{chen2020inception,
  AUTHOR = {Chen, Wanpei and Qiao, Yanting and Li, Yujie},
  PUBLISHER = {Springer},
  DATE = {2020},
  JOURNALTITLE = {J. Ambient Intell. Humaniz. Comput.},
  PAGES = {1--7},
  TITLE = {Inception-{{SSD}}: {{An}} Improved Single Shot Detector for Vehicle Detection},
}

@INPROCEEDINGS{chen2007real,
  AUTHOR = {Chen, Yu Ming and Dong, Liang and Oh, Jun-Seok},
  ORGANIZATION = {IEEE},
  BOOKTITLE = {2007 {{IEEE Wirel}}. {{Commun}}. {{Netw}}. {{Conf}}.},
  DATE = {2007},
  DOI = {10.1109/wcnc.2007.485},
  TITLE = {Real-Time Video Relay for Uav Traffic Surveillance Systems through Available Communication Networks},
}

@INPROCEEDINGS{chu2017multi,
  AUTHOR = {Chu, Xiao and Yang, Wei and Ouyang, Wanli and Ma, Cheng and Yuille, Alan L and Wang, Xiaogang},
  BOOKTITLE = {Proceedings of the {{IEEE}} Conference on Computer Vision and Pattern Recognition},
  DATE = {2017},
  TITLE = {Multi-Context Attention for Human Pose Estimation},
}

@INPROCEEDINGS{bnlstm,
  AUTHOR = {Cooijmans, Tim and Ballas, Nicolas and Laurent, César and Gülçehre, Çaglar and Courville, Aaron C.},
  BOOKTITLE = {5th {{Int}}. {{Conf}}. {{Learn}}. {{Represent}}. {{ICLR}} 2017 {{Toulon Fr}}. {{April}} 24-26 2017 {{Conf}}. {{Track Proc}}.},
  DATE = {2017},
  TITLE = {Recurrent Batch Normalization},
}

@INPROCEEDINGS{dike2018unmanned,
  AUTHOR = {Dike, Happiness Ugochi and Wu, Qingtian and Zhou, Yimin and Liang, Gong},
  ORGANIZATION = {IEEE},
  BOOKTITLE = {2018 {{IEEE}} International Conference on Robotics and Biomimetics ({{ROBIO}})},
  DATE = {2018},
  DOI = {10/gnbp6s},
  TITLE = {Unmanned Aerial Vehicle ({{UAV}}) Based Running Person Detection from a Real-Time Moving Camera},
}

@ARTICLE{douklias2022design,
  AUTHOR = {Douklias, Athanasios and Karagiannidis, Lazaros and Misichroni, Fay and Amditis, Angelos},
  PUBLISHER = {MDPI},
  DATE = {2022},
  DOI = {10.3390/s22052049},
  JOURNALTITLE = {Sensors},
  NUMBER = {5},
  PAGES = {2049},
  TITLE = {Design and Implementation of a {{UAV-Based}} Airborne Computing Platform for Computer Vision and Machine Learning Applications},
  VOLUME = {22},
}

@INPROCEEDINGS{hager2016combining,
  AUTHOR = {Häger, Gustav and Bhat, Goutam and Danelljan, Martin and Khan, Fahad Shahbaz and Felsberg, Michael and Rudl, Piotr and Doherty, Patrick},
  ORGANIZATION = {Springer},
  BOOKTITLE = {International Symposium on Visual Computing},
  DATE = {2016},
  TITLE = {Combining Visual Tracking and Person Detection for Long Term Tracking on a Uav},
}

@ARTICLE{hildmann2019using,
  AUTHOR = {Hildmann, Hanno and Kovacs, Ernö},
  PUBLISHER = {Multidisciplinary Digital Publishing Institute},
  DATE = {2019},
  DOI = {10.3390/drones3030059},
  JOURNALTITLE = {Drones},
  NUMBER = {3},
  PAGES = {59},
  TITLE = {Using Unmanned Aerial Vehicles ({{UAVs}}) as Mobile Sensing Platforms ({{MSPs}}) for Disaster Response, Civil Security and Public Safety},
  VOLUME = {3},
}

@INPROCEEDINGS{hong2016learning,
  AUTHOR = {Hong, Seunghoon and Oh, Junhyuk and Lee, Honglak and Han, Bohyung},
  BOOKTITLE = {Proceedings of the {{IEEE}} Conference on Computer Vision and Pattern Recognition},
  DATE = {2016},
  TITLE = {Learning Transferrable Knowledge for Semantic Segmentation with Deep Convolutional Neural Network},
}

@ARTICLE{sensorsMLforDiabetes,
  AUTHOR = {Hu, Hansel and Lai, Tin and Farid, Farnaz},
  URL = {https://www.mdpi.com/1424-8220/22/16/6155},
  DATE = {2022},
  DOI = {10.3390/s22166155},
  ISSN = {1424-8220},
  JOURNALTITLE = {Applications of Body Worn Sensors and Wearables, \emph{Special Issue of} Sensors},
  NUMBER = {16},
  PAGES = {6155},
  TITLE = {Feasibility Study of Constructing a Screening Tool for Adolescent Diabetes Detection Applying Machine Learning Methods},
  VOLUME = {22},
}

@ARTICLE{huang2019lightweight,
  AUTHOR = {Huang, Kai and Liu, Ximeng and Fu, Shaojing and Guo, Deke and Xu, Ming},
  PUBLISHER = {IEEE},
  DATE = {2019},
  JOURNALTITLE = {IEEE Transactions on Dependable and Secure Computing},
  TITLE = {A Lightweight Privacy-Preserving {{CNN}} Feature Extraction Framework for Mobile Sensing},
}

@ARTICLE{hulens2015choose,
  AUTHOR = {Hulens, Dries and Goedemé, Toon and Verbeke, Jon},
  DATE = {2015},
  JOURNALTITLE = {Proceedings VISAPP 2015},
  PAGES = {1--10},
  TITLE = {How to Choose the Best Embedded Processing Platform for On-Board {{UAV}} Image Processing?},
}

@INPROCEEDINGS{ibrahim2018hierarchical,
  AUTHOR = {Ibrahim, Mostafa S and Mori, Greg},
  BOOKTITLE = {Proceedings of the {{European}} Conference on Computer Vision ({{ECCV}})},
  DATE = {2018},
  DOI = {10.1007/978-3-030-01219-9_44},
  TITLE = {Hierarchical Relational Networks for Group Activity Recognition and Retrieval},
}

@ARTICLE{ji20133d,
  AUTHOR = {Ji, Shuiwang and Xu, Wei and Yang, Ming and Yu, Kai},
  PUBLISHER = {IEEE},
  DATE = {2013},
  JOURNALTITLE = {IEEE transactions on pattern analysis and machine intelligence},
  NUMBER = {1},
  PAGES = {221--231},
  TITLE = {{{3D}} Convolutional Neural Networks for Human Action Recognition},
  VOLUME = {35},
}

@ARTICLE{al2018survey,
  OPTIONS = {useprefix=1},
  AUTHOR = {Al-Kaff, Abdulla and Martin, David and Garcia, Fernando and de la Escalera, Arturo and Armingol, Jose Maria},
  PUBLISHER = {Elsevier},
  DATE = {2018},
  DOI = {10.1016/j.eswa.2017.09.033},
  JOURNALTITLE = {Expert Systems With Applications},
  PAGES = {447--463},
  SHORTJOURNAL = {Expert Syst. Appl.},
  TITLE = {Survey of Computer Vision Algorithms and Applications for Unmanned Aerial Vehicles},
  VOLUME = {92},
}

@ARTICLE{kakaletsis2021computer,
  AUTHOR = {Kakaletsis, Efstratios and Symeonidis, Charalampos and Tzelepi, Maria and Mademlis, Ioannis and Tefas, Anastasios and Nikolaidis, Nikos and Pitas, Ioannis},
  PUBLISHER = {ACM New York, NY},
  DATE = {2021},
  DOI = {10.1145/3472288},
  JOURNALTITLE = {ACM Comput. Surv. CSUR},
  NUMBER = {9},
  PAGES = {1--37},
  TITLE = {Computer Vision for Autonomous {{UAV}} Flight Safety: {{An}} Overview and a Vision-Based Safe Landing Pipeline Example},
  VOLUME = {54},
}

@ARTICLE{karaca2018potential,
  AUTHOR = {Karaca, Yunus and Cicek, Mustafa and Tatli, Ozgur and Sahin, Aynur and Pasli, Sinan and Beser, Muhammed Fatih and Turedi, Suleyman},
  PUBLISHER = {Elsevier},
  DATE = {2018},
  DOI = {10.1016/j.ajem.2017.09.025},
  JOURNALTITLE = {American Journal of Emergency Medicine},
  NUMBER = {4},
  PAGES = {583--588},
  SHORTJOURNAL = {Am. J. Emerg. Med.},
  TITLE = {The Potential Use of Unmanned Aircraft Systems (Drones) in Mountain Search and Rescue Operations},
  VOLUME = {36},
}

@ARTICLE{kraft2021autonomous,
  AUTHOR = {Kraft, Marek and Piechocki, Mateusz and Ptak, Bartosz and Walas, Krzysztof},
  PUBLISHER = {Multidisciplinary Digital Publishing Institute},
  DATE = {2021},
  DOI = {10/gjkt3z},
  JOURNALTITLE = {Remote Sensing},
  NUMBER = {5},
  PAGES = {965},
  SHORTJOURNAL = {Remote Sens.},
  TITLE = {Autonomous, Onboard Vision-Based Trash and Litter Detection in Low Altitude Aerial Images Collected by an Unmanned Aerial Vehicle},
  VOLUME = {13},
}

@ARTICLE{lai2022slamreview,
  AUTHOR = {Lai, Tin},
  PUBLISHER = {MDPI},
  DATE = {2022},
  JOURNALTITLE = {Sensors},
  NUMBER = {19},
  PAGES = {7265},
  TITLE = {A Review on Visual-{{SLAM}}: {{Advancements}} from Geometric Modelling to Learning-Based Semantic Scene Understanding Using Multi-Modal Sensor Fusion},
  VOLUME = {22},
}

@ARTICLE{lai2021rrf,
  AUTHOR = {Lai, Tin and Ramos, Fabio},
  DATE = {2022},
  DOI = {10.1109/LRA.2021.3132985},
  JOURNALTITLE = {IEEE Robotics and Automation Letters (RA-L)},
  NUMBER = {2},
  PAGES = {1111--1117},
  TITLE = {Adaptively Exploits Local Structure with Generalised Multi-Trees Motion Planning},
  VOLUME = {7},
}

@ARTICLE{lai2021kinoSBP,
  AUTHOR = {Lai, Tin and Zhi, Weiming and Hermans, Tucker and Ramos, Fabio},
  DATE = {2022},
  JOURNALTITLE = {Computing Research Repository (CoRR)},
  TITLE = {{{L4KDE}}: {{Learning}} for {{KinoDynamic}} Tree Expansion},
}

@INPROCEEDINGS{liu2016ssd,
  AUTHOR = {Liu, Wei and Anguelov, Dragomir and Erhan, Dumitru and Szegedy, Christian and Reed, Scott and Fu, Cheng-Yang and Berg, Alexander C},
  ORGANIZATION = {Springer},
  BOOKTITLE = {Eur. {{Conf}}. {{Comput}}. {{Vis}}.},
  DATE = {2016},
  DOI = {10.1007/978-3-319-46448-0_2},
  TITLE = {Ssd: {{Single}} Shot Multibox Detector},
}

@ARTICLE{merino2006cooperative,
  AUTHOR = {Merino, Luis and Caballero, Fernando and Martínez-de Dios, J Ramiro and Ferruz, Joaquín and Ollero, Aníbal},
  PUBLISHER = {Wiley Online Library},
  DATE = {2006},
  JOURNALTITLE = {Journal of Field Robotics},
  NUMBER = {3-4},
  PAGES = {165--184},
  TITLE = {A Cooperative Perception System for Multiple {{UAVs}}: {{Application}} to Automatic Detection of Forest Fires},
  VOLUME = {23},
}

@INPROCEEDINGS{neubeck2006efficient,
  AUTHOR = {Neubeck, Alexander and Van Gool, Luc},
  ORGANIZATION = {IEEE},
  BOOKTITLE = {18th {{Int}}. {{Conf}}. {{Pattern Recognit}}. {{ICPR06}}},
  DATE = {2006},
  DOI = {10.1109/icpr.2006.479},
  TITLE = {Efficient Non-Maximum Suppression},
}

@INPROCEEDINGS{ouyang2013joint,
  AUTHOR = {Ouyang, Wanli and Wang, Xiaogang},
  BOOKTITLE = {Proceedings of the {{IEEE}} International Conference on Computer Vision},
  DATE = {2013},
  DOI = {10.1109/ICCV.2013.257},
  TITLE = {Joint Deep Learning for Pedestrian Detection},
}

@INPROCEEDINGS{ramanathan2016detecting,
  AUTHOR = {Ramanathan, Vignesh and Huang, Jonathan and Abu-El-Haija, Sami and Gorban, Alexander and Murphy, Kevin and Fei-Fei, Li},
  BOOKTITLE = {Proceedings of the {{IEEE}} Conference on Computer Vision and Pattern Recognition},
  DATE = {2016},
  DOI = {10.1109/CVPR.2016.332},
  TITLE = {Detecting Events and Key Actors in Multi-Person Videos},
}

@INPROCEEDINGS{redmon2016yolo,
  AUTHOR = {Redmon, Joseph and Divvala, Santosh and Girshick, Ross and Farhadi, Ali},
  BOOKTITLE = {The {{IEEE}} Conference on Computer Vision and Pattern Recognition ({{CVPR}})},
  DATE = {2016},
  DOI = {10.1109/cvpr.2016.91},
  TITLE = {You Only Look Once: {{Unified}}, Real-Time Object Detection},
}

@ONLINE{redmon2018yolov3,
  AUTHOR = {Redmon, Joseph and Farhadi, Ali},
  DATE = {2018},
  EPRINT = {1804.02767},
  EPRINTTYPE = {arxiv},
  TITLE = {Yolov3: {{An}} Incremental Improvement},
}

@ARTICLE{ren2015faster,
  AUTHOR = {Ren, Shaoqing and He, Kaiming and Girshick, Ross and Sun, Jian},
  DATE = {2015},
  JOURNALTITLE = {Advances in neural information processing systems},
  PAGES = {91--99},
  SHORTJOURNAL = {Adv. Neural Inf. Process. Syst.},
  TITLE = {Faster {{R-cnn}}: {{Towards}} Real-Time Object Detection with Region Proposal Networks},
  VOLUME = {28},
}

@INPROCEEDINGS{sahingoz2013mobile,
  AUTHOR = {Sahingoz, Ozgur Koray},
  ORGANIZATION = {IEEE},
  BOOKTITLE = {2013 International Conference on Unmanned Aircraft Systems ({{ICUAS}})},
  DATE = {2013},
  TITLE = {Mobile Networking with {{UAVs}}: {{Opportunities}} and Challenges},
}

@INPROCEEDINGS{sambolek2020person,
  AUTHOR = {Sambolek, Sasa and Ivasic-Kos, Marina},
  ORGANIZATION = {IEEE},
  BOOKTITLE = {2020 5th International Conference on Smart and Sustainable Technologies ({{SpliTech}})},
  DATE = {2020},
  DOI = {10/gnbp6v},
  TITLE = {Person Detection in Drone Imagery},
}

@INPROCEEDINGS{sandler2018mobilenetv2,
  AUTHOR = {Sandler, Mark and Howard, Andrew and Zhu, Menglong and Zhmoginov, Andrey and Chen, Liang-Chieh},
  BOOKTITLE = {Proceedings of the {{IEEE}} Conference on Computer Vision and Pattern Recognition},
  DATE = {2018},
  DOI = {10.1109/CVPR.2018.00474},
  TITLE = {Mobilenetv2: {{Inverted}} Residuals and Linear Bottlenecks},
}

@ARTICLE{schulte2017_AnalComb,
  AUTHOR = {Schulte, S. and Hillen, F. and Prinz, T.},
  URL = {https://www.int-arch-photogramm-remote-sens-spatial-inf-sci.net/XLII-2-W6/347/2017/},
  DATE = {2017-08-24},
  DOI = {10.5194/isprs-archives-xlii-2-w6-347-2017},
  ISSN = {2194-9034},
  JOURNALTITLE = {ISPRS - International Archives of the Photogrammetry, Remote Sensing and Spatial Information Sciences},
  LANGID = {english},
  PAGES = {347--354},
  SHORTJOURNAL = {Int. Arch. Photogramm. Remote Sens. Spatial Inf. Sci.},
  SHORTTITLE = {Analysis of Combined Uav-Based Rgb and Thermal Remote Sensing Data},
  TITLE = {Analysis of Combined Uav-Based Rgb and Thermal Remote Sensing Data: {{A}} New Approach to Crowd Monitoring},
  URLDATE = {2020-08-18},
  VOLUME = {XLII-2/W6},
}

@INPROCEEDINGS{sozykin2018multi,
  AUTHOR = {Sozykin, Konstantin and Protasov, Stanislav and Khan, Adil and Hussain, Rasheed and Lee, Jooyoung},
  ORGANIZATION = {IEEE},
  BOOKTITLE = {2018 19th {{IEEE}}/{{ACIS}} International Conference on Software Engineering, Artificial Intelligence, Networking and {{Parallel}}/{{Distributed}} Computing ({{SNPD}})},
  DATE = {2018},
  DOI = {10/gnbp6t},
  TITLE = {Multi-Label Class-Imbalanced Action Recognition in Hockey Videos via {{3D}} Convolutional Neural Networks},
}

@INPROCEEDINGS{stewart2016end,
  AUTHOR = {Stewart, Russell and Andriluka, Mykhaylo and Ng, Andrew Y},
  BOOKTITLE = {Proceedings of the {{IEEE}} Conference on Computer Vision and Pattern Recognition},
  DATE = {2016},
  DOI = {10.1109/CVPR.2016.255},
  TITLE = {End-to-End People Detection in Crowded Scenes},
}

@INPROCEEDINGS{szegedy2016rethinking,
  AUTHOR = {Szegedy, Christian and Vanhoucke, Vincent and Ioffe, Sergey and Shlens, Jon and Wojna, Zbigniew},
  BOOKTITLE = {Proceedings of the {{IEEE}} Conference on Computer Vision and Pattern Recognition},
  DATE = {2016},
  DOI = {10.1109/CVPR.2016.308},
  TITLE = {Rethinking the Inception Architecture for Computer Vision},
}

@INPROCEEDINGS{tian2015deep,
  AUTHOR = {Tian, Yonglong and Luo, Ping and Wang, Xiaogang and Tang, Xiaoou},
  BOOKTITLE = {Proceedings of the {{IEEE}} International Conference on Computer Vision},
  DATE = {2015},
  DOI = {10.1109/ICCV.2015.221},
  TITLE = {Deep Learning Strong Parts for Pedestrian Detection},
}

@INPROCEEDINGS{vega2015resilient,
  AUTHOR = {Vega, Augusto and Lin, Chung-Ching and Swaminathan, Karthik and Buyuktosunoglu, Alper and Pankanti, Sharathchandra and Bose, Pradip},
  ORGANIZATION = {IEEE},
  BOOKTITLE = {2015 33rd {{IEEE}} International Conference on Computer Design ({{ICCD}})},
  DATE = {2015},
  TITLE = {Resilient, {{UAV-embedded}} Real-Time Computing},
}

@INPROCEEDINGS{wang2016differential,
  AUTHOR = {Wang, Chu and Siddiqi, Kaleem},
  BOOKTITLE = {Proceedings of the {{IEEE}} Conference on Computer Vision and Pattern Recognition Workshops},
  DATE = {2016},
  TITLE = {Differential Geometry Boosts Convolutional Neural Networks for Object Detection},
}

@INPROCEEDINGS{wang2018fast,
  AUTHOR = {Wang, Xiaoliang and Cheng, Peng and Liu, Xinchuan and Uzochukwu, Benedict},
  ORGANIZATION = {IEEE},
  BOOKTITLE = {{{IECON}} 2018-44th Annual Conference of the {{IEEE}} Industrial Electronics Society},
  DATE = {2018},
  DOI = {10/gnbp6q},
  TITLE = {Fast and Accurate, Convolutional Neural Network Based Approach for Object Detection from {{UAV}}},
}

@INPROCEEDINGS{wang2009hog,
  AUTHOR = {Wang, Xiaoyu and Han, Tony X and Yan, Shuicheng},
  ORGANIZATION = {IEEE},
  BOOKTITLE = {2009 {{IEEE}} 12th {{Int}}. {{Conf}}. {{Comput}}. {{Vis}}.},
  DATE = {2009},
  DOI = {10.1109/ICCV.2009.5459207},
  TITLE = {An {{HOG-LBP}} Human Detector with Partial Occlusion Handling},
}

@ARTICLE{forexNonStationaryTimeSeries,
  AUTHOR = {Wang, Xipei and Zhang, Haoyu and Zhang, Yuanbo and Wang, Meng and Song, Jiarui and Lai, Tin and Khushi, Matloob},
  DATE = {2022},
  DOI = {10.1109/TAI.2021.3130529},
  JOURNALTITLE = {IEEE Transactions on Artificial Intelligence (TAI)},
  NUMBER = {5},
  PAGES = {778--787},
  TITLE = {Learning Non-Stationary Time-Series with Dynamic Pattern Extractions},
  VOLUME = {3},
}

@INPROCEEDINGS{xu2017r,
  AUTHOR = {Xu, Huijuan and Das, Abir and Saenko, Kate},
  BOOKTITLE = {Proceedings of the {{IEEE}} International Conference on Computer Vision},
  DATE = {2017},
  TITLE = {R-{{C3d}}: {{Region}} Convolutional 3d Network for Temporal Activity Detection},
}

@ARTICLE{xu2022waterSedimentML,
  AUTHOR = {Xu, Xiaoting and Lai, Tin and Jahan, Sayka and Farid, Farnaz and Bello, Abubakar},
  PUBLISHER = {MDPI},
  DATE = {2022},
  JOURNALTITLE = {Future Internet},
  NUMBER = {11},
  PAGES = {324},
  TITLE = {A Machine Learning Predictive Model to Detect Water Quality and Pollution},
  VOLUME = {14},
}

@ARTICLE{xu2017deformable,
  AUTHOR = {Xu, Zhaozhuo and Xu, Xin and Wang, Lei and Yang, Rui and Pu, Fangling},
  PUBLISHER = {Multidisciplinary Digital Publishing Institute},
  DATE = {2017},
  DOI = {10.3390/rs9121312},
  JOURNALTITLE = {Remote Sens.},
  NUMBER = {12},
  PAGES = {1312},
  TITLE = {Deformable Convnet with Aspect Ratio Constrained Nms for Object Detection in Remote Sensing Imagery},
  VOLUME = {9},
}

@ARTICLE{yang2019effective,
  AUTHOR = {Yang, Jianxiu and Xie, Xuemei and Yang, Wenzhe},
  PUBLISHER = {IEEE},
  DATE = {2019},
  JOURNALTITLE = {IEEE access : practical innovations, open solutions},
  PAGES = {85042--85054},
  SHORTJOURNAL = {IEEE Access},
  TITLE = {Effective Contexts for {{UAV}} Vehicle Detection},
  VOLUME = {7},
}

@ARTICLE{zhang2020object,
  AUTHOR = {Zhang, Ruiqian and Shao, Zhenfeng and Huang, Xiao and Wang, Jiaming and Li, Deren},
  PUBLISHER = {Multidisciplinary Digital Publishing Institute},
  DATE = {2020},
  DOI = {10/gh2j82},
  JOURNALTITLE = {Remote Sensing},
  NUMBER = {19},
  PAGES = {3140},
  SHORTJOURNAL = {Remote Sens.},
  TITLE = {Object Detection in {{UAV}} Images via Global Density Fused Convolutional Network},
  VOLUME = {12},
}

@INPROCEEDINGS{zheng2017learning,
  AUTHOR = {Zheng, Heliang and Fu, Jianlong and Mei, Tao and Luo, Jiebo},
  BOOKTITLE = {Proceedings of the {{IEEE}} International Conference on Computer Vision},
  DATE = {2017},
  DOI = {10.1109/ICCV.2017.557},
  TITLE = {Learning Multi-Attention Convolutional Neural Network for Fine-Grained Image Recognition},
}

@ONLINE{zheng2016good,
  AUTHOR = {Zheng, Liang and Zhao, Yali and Wang, Shengjin and Wang, Jingdong and Tian, Qi},
  DATE = {2016},
  EPRINT = {1604.00133},
  EPRINTTYPE = {arxiv},
  TITLE = {Good Practice in {{CNN}} Feature Transfer},
}

\end{document}